%% file: main.tex
\documentclass[letterpaper, 10 pt, conference]{ieeeconf}  

\let\labelindent\relax
\usepackage{enumitem}

\IEEEoverridecommandlockouts                              

\usepackage{graphics} 
\usepackage{graphicx}
\usepackage{times} 
\usepackage{amsmath} 
\usepackage{amssymb}  
\usepackage{booktabs}
\usepackage{tikz}
\usepackage{enumitem}
\usepackage{pifont}
\usepackage{url}
\usepackage{caption}

\usepackage{cite}

\newcommand{\cmark}{\ding{51}}
\newcommand{\xmark}{\ding{55}}

\title{\LARGE \bf
Sceniris: A Fast Procedural Scene Generation Framework
}

\author{Jinghuan Shang$^{1}$, Harsh Patel$^{2}$, Ran Gong$^{1}$, Karl Schmeckpeper$^{1}$
\thanks{$^{1}$Robotics and AI Institute, Boston, MA, USA.
        {\tt\small \{jshang, rgong, kschmeckpeper\}@rai-inst.com}}%
\thanks{$^{2}$University of Waterloo, Ontario, Canada. Work done during the internship at Robotics and AI Institute.
        {\tt\small h329pate@uwaterloo.ca}}%
}
\IEEEaftertitletext{
\input{figures/teaser}
}

\begin{document}
\maketitle
\thispagestyle{empty}
\pagestyle{empty}
\begin{abstract}
    Synthetic 3D scenes are essential for developing Physical AI and generative models. Existing procedural generation methods often have low output throughput, creating a significant bottleneck in scaling up dataset creation. In this work, we introduce Sceniris, a highly efficient procedural scene generation framework for rapidly generating large-scale, collision-free scene variations. Sceniris also provides an optional robot reachability check, providing manipulation-feasible scenes for robot tasks. Sceniris is designed for maximum efficiency by addressing the primary performance limitations of the prior method, Scene Synthesizer. Leveraging batch sampling and faster collision checking in cuRobo, Sceniris achieves at least 234x speed-up over Scene Synthesizer. Sceniris also expands the object-wise spatial relationships available in prior work to support diverse scene requirements. Our code is available at \url{https://github.com/rai-inst/sceniris}
\end{abstract}


\section{INTRODUCTION}
Data-driven approaches are one of the most effective methods to develop AI when data are sufficient. Internet-scale language and image data have successfully spawned many Large Language Models (LLMs)~\cite{achiam2023gpt,touvron2023llama,bai2023qwen,liu2024deepseek}, Vision Foundation Models (VFMs)~\cite{dosovitskiy2021an,oquab2023dinov2,simeoni2025dinov3,shang2024theia,ranzinger2024radio,heinrich2025radiov2}, and Vision-Language Models (VLMs)~\cite{alayrac2022flamingo,tschannen2025siglip,zhai2023sigmoid,liu2023visual,radford2021learning}. Moving forward from word and pixels, the community is developing AI that perceives, generates copies of, or interacts with the physical world where the existing dataset is limited~\cite{grauman2022ego4d,damen2018scaling,khazatsky2024droid,o2024open,zhang2024vla}. Collecting sufficient real-world data is also extremely time-consuming~\cite{khazatsky2024droid,black2024pi0,galaxeag0}. As a result, synthetic environments and data are another important source of data~\cite{scenesynthesizer,procthor,robotwin2,robomimic,robocasa}.

In this work, we focus on providing synthetic scenes for physical AI, like simulated environments for robots~\cite{gensim,gensim2,robogen,robocasa}, training generative AI~\cite{pfaff2025_steerable_scene_generation, li2024dreamscene}, and 3D perception tasks~\cite{peng2023openscene,hou2021exploring,chen2023clip2scene} at a large scale. We aim to generate random object poses that follow the scene layout configuration provided by the user and ensure the scene is collision-free. Existing works on generating synthetic scenes are either procedural-based~\cite{izatt2020generative,scenesynthesizer,procthor} or learning-based~\cite{li2024dreamscene,chen2023scenedreamer,shriram2024realmdreamer}. Learning-based methods, which are mostly generative AI, require a dataset that is usually generated by procedural-based methods. However, the procedural-based approach often takes a long time to produce a dataset at an acceptable scale~\cite{pfaff2025_steerable_scene_generation,izatt2020generative}. The efficiency of the procedural-based approach becomes a key bottleneck for scaling up.

We introduce Sceniris, a fast procedural scene generation framework that generates a large number of collision-free variations of scenes in a very short time. Sceniris also provides robot reachability check for cases generating scenes that guarantee specified objects are reachable by the robot, which is essential for embodied AI, and none of the existing scene generation framework supports this. Sceniris is designed with the top priority on efficiency. We propose the idea of parallelizing the sampling process and collision checking, two major bottlenecks in the existing procedural generation method, to achieve the best efficiency. We demonstrate our parallelization based on a prior work on procedural scene generation, Scene Synthesizer~\cite{scenesynthesizer}. Specifically, we leverage faster APIs in geometry libraries~\cite{shapely2007} and use batch matrix computations to speed up sampling, and we use cuRobo~\cite{curobo} to speed up collision checking. In addition, we add more spatial relationships that are not available in scene synthesizer~\cite{scenesynthesizer}. With all the improvements, Sceniris achieves more than 234$\times$ speed-up than scene synthesizer~\cite{scenesynthesizer}, showing its strong ability for scaling up the procedural scene generation.

\vspace{-6pt}
\section{RELATED WORK}
\vspace{-4pt}
\subsection{Applications that Require 3D Scenes}
The demand for diverse 3D scenes has grown significantly across various domains. In robotics, diverse scenes are required in simulation~\cite{mujoco,isaacsim,isaaclab,pybullet,metaworld,robosuite2020,coppeliasim,geng2025roboverse,carla,habitat3,habitat2,habitat1} to train policies~\cite{rlbench,libero,calvin,vima,simplerenv,robocasa,arnold,robotwin1} and collect data~\cite{robomimic}. Randomizing the object poses within a scene is also important to improve the generalizability of the policy and increase the diversity of the data~\cite{rlbench,vima,robocasa,arnold}. Moreover, digital twins~\cite{lumer2021towardsadigital} and digital cousins~\cite{dai2024acdc} aim at minimizing sim-to-real gap, where higher quality 3D scenes are more demanded. 
Generative simulations~\cite{procthor,robogen,gensim,gensim2,robotwin2}, which generate the scene on-the-fly, require a reliable approach to generate object layouts. 
Training generative models for 3D scenes also requires a huge amount of data~\cite{ocal2024sceneteller,chen2023scenedreamer,zhang20243d,zhang2024text2nerf,li2024dreamscene,lin2024genusd,shriram2024realmdreamer,pfaff2025_steerable_scene_generation}, where procedure scene generation~\cite{chang2015text,izatt2020generative,raistrick2023infinite,raistrick2024infinigen,joshi2025infinigen,scenesynthesizer} methods are often required to produce them.
3D perception tasks can also benefit from a large 3D scene dataset. Therefore, we believe procedural scene generation methods are a good fit for all these applications, and the methods that have higher output throughput will greatly support the high data demand.

\input{figures/main_steps}
\vspace{-4pt}
\subsection{Scene Generation Approaches}
Procedural-based and learning-based scene generation approaches are the two ways to obtain synthesized scenes for the application above. Procedural-based methods often involve sampling procedures and require a schema or a configuration to decide the scene layout. Infinigen~\cite{raistrick2023infinite,raistrick2024infinigen} can generate high-quality multi-room scale indoor scenes using Monte-Carlo Metropolis-Hastings sampling. ProcTHOR~\cite{procthor} also generates indoor scenes and places objects in a collision-free manner. Scene Synthesizer~\cite{scenesynthesizer} can compose both procedural-based objects and scenes like kitchens, and the object layout is generated by rejection sampling with collision checking. Spatial scene grammars~\cite{izatt2020generative} generate the scenes on CPU. 
There are also algorithms embedded in simulation frameworks like GenSim~\cite{gensim}, GenSim2~\cite{gensim2}, RoboGen~\cite{robogen}, and Robocasa~\cite{robocasa}, to compose the scene and randomize the object poses, but usually the range of randomizing the poses is limited.
UE5 PCG~\cite{ue5pcg} plugin is a scene generation tool in the game engine to generate a random scene. A more comprehensive comparison across procedural-based scene generation is available in Table~\ref{tab:sg_comparision}.

However, the existing methods do not scale to generating a large batch of scenes in a short time. For example, Infinigen takes up to 10 minutes to generate a living room, and Scene Synthesizer generates one table-top scene in about 10 seconds. Under this throughput, generating a large-scale dataset for 3D scenes or performing object pose randomization is difficult.
Learning-based methods, like steerable scene generation~\cite{pfaff2025_steerable_scene_generation}, may also have a similar throughput and require a huge dataset to train.
To this end, our objective is to generate 100-1000 scenes within a second on average, providing strong support for data-hungry scenarios.


\section{Preliminary: Scene Synthesizer}
Scene Synthesizer~\cite{scenesynthesizer} is a single-threaded scene generation framework. It procedurally places objects in the scene, as well as generates some kitchen or office objects. In this work, we mainly discuss the object placement part, which is representative of existing single-threaded procedural scene generation methods. Our objective is to design a faster procedural scene generation framework by parallelizing the key steps, and Scene Synthesizer~\cite{scenesynthesizer} is a great base approach that can demonstrate our improvements. Below, we go through the Scene Synthesizer's scene generation process and key components, and identify components to improve.

\paragraph{Main algorithm} Scene Synthesizer~\cite{scenesynthesizer} uses rejection sampling as the main generation logic. When adding a new object to the scene, it first samples \textbf{one} possible pose for the object. The position consists of a 2D position sampled in a previously marked 2D surface plus a collision-avoiding z coordinate, or a 3D position directly sampled from a 3D volume. The orientation is sampled from pre-defined rules (e.g., random rotation along the z-axis) or stable poses computed from the object mesh. Then, the collision checker verifies whether the object can be placed at the sampled pose without any collisions. Retry on an object happens if any collision is detected. 
The entire main logic runs on a single main thread. We observe an opportunity to speed up the algorithm by parallelization. 

\paragraph{Scene representation} Scene Synthesizer uses Trimesh to represent the scene as a scene graph, where a node is an object or an object part, and a directed edge is the transform from the source node to the destination node.
The transform is stored in the 4$\times$4 homogeneous matrix. As a result, Scene Synthesizer can represent and work on one scene instance at any time. We observe that there is room for improvement through the representation of a batch of scenes.

\paragraph{Collision check} Scene Synthesizer uses trimesh's collision manager to detect collision. Trimesh uses Flexible Collision Library (FCL) as the underlying algorithm. FCL runs on the CPU entirely.

\paragraph{Position sampling} For an object to be placed on a surface, which is the most common case in scene generation, Scene Synthesizer uses \texttt{shapely}~\cite{shapely2007} to sample one 2D position from the pre-computed surface represented by a polygon. The polygon is called the support polygon. 
It first randomly samples a 2D point within the polygon and attaches a non-zero offset on the z-axis to avoid collision.
Though a vectorized (i.e., high efficiency) API is used during sampling, the actual use is sampling only one position from the polygon at a time, which is not efficient. 

\vspace{-3pt}
\section{METHOD}
\input{tables/sg_comparision}
Sceniris is developed based on Scene Synthesizer~\cite{scenesynthesizer}, and our top priority is improving its efficiency. Our key design idea is to sample a large number of scenes in a batch that leverages the parallelization capability of computing libraries. The main bottlenecks in Scene Synthesizer are pose sampling and collision checking, as we discussed above. To this end, we improve the sampling by batch and improve the collision checking using cuRobo. In addition, to support more spatial relationships for procedural generation, we extend the existing spatial relationships and sample them efficiently. We also add a reachability feature to support robot manipulation scene generation. We describe these main improvements in the following subsections. 

\vspace{-5pt}
\subsection{System Overview} The main scene generation procedure is shown in Figure~\ref{fig:main_steps}. Sceniris first initializes the system, including loading the assets, determining the order of object placement, preparing the pose samplers (including spatial relationships), and preparing the collision checker. Then, each object will be placed into the scene. For each object, Sceniris samples its poses, performs collision checking, and optionally retries limited times for environments that fail the check. The scene generation procedure ends once all objects are successfully placed or after the maximum number of retries. 

\vspace{-5pt}
\subsection{Batched Scene Representation}
The core supporting modification is batched scene representation -- storing a batch of scene instances in the scene graph. We store a batch of $N$ transform matrices in edges that represent the $N$ instances of the spatial relationship between two nodes. For example, if we are generating 8 different scenes where an apple is randomly placed on the table, the (table, apple) edge stores an $(8,4,4)$ tensor representing the 8 transform matrices. Based on this batched scene representation, we also implemented a batched forward kinematics function, allowing us to modify the poses of articulated object parts in parallel. With the batched scene representation, Sceniris can support more improvements that require batch operations.

\vspace{-5pt}
\subsection{Batch Sampling and Caching}
The original Scene Synthesizer heavily relies on sampling 2D points from a polygon, but only one point is sampled at a time. 
We sample a batch of points at a time instead. However, calling sampling multiple times still has overhead. Instead of sampling in an ad-hoc way, we first sample a larger batch than required and then cache the batch in a queue. When the cached samples run out, it samples another batch. With caching, the system can continuously generate more batches of scene instances with less overhead.

The batch sampling method above covers the case of one support polygon. Considering the case that we want to randomly place an apple on the table in $N$ scene instances, and the table is also randomly placed on the ground, the support polygons for the apple are the same table surface with different world poses. For a case like this, instead of sampling points from each polygon, we sample $N$ (or more using cache) points from the canonicalized polygon (usually the first scene instance), and transform each point using the world pose of its corresponding support polygon for each scene instance. This approach significantly reduces overhead for iterating through scene instances. However, it is not always the case that the support polygon can be affine-transformed from the first scene instance. This usually happens when we customize complex spatial constraints introduced in Section~\ref{sec:spatial_relationship}. For these cases, the system falls back to sample points from each polygon.

\vspace{-5pt}
\subsection{Collision Checking}
Once a batch of poses is sampled, they should be verified by the collision checker to filter out non-collision-free placements. We use cuRobo~\cite{curobo} to perform collision checking for a batch of scenes on GPU. To minimize the overhead, we pre-cached all the meshes and the world configurations that cuRobo requires during the initialization stage, which will be executed only once. We only modify the meshes enabled/disabled state and the transforms through cuRobo's collision managing instance. During retry, the collision checking is performed only on those scenes that require retry, further reducing the overhead compared to checking all the scenes. 

\subsection{Reachability}
Understanding a manipulator’s workspace is essential for many tasks, as it plays a key role in configuring a scene for feasible task execution. To address this, we utilize RM4D \cite{rudorfer2024rm4d} — a reachability map that uses a single 4D data structure to support both forward and inverse queries without sacrificing accuracy. To further enhance performance, we batch process queries and transfer the data structure to the GPU, significantly increasing throughput and efficiency. The reachability check feature can be optionally turned on for objects or object parts, ensuring the object or parts are reachable. Given a (hypothesized) robot position in the scene configuration and the objects (parts) that require the reachability check, the checker rejects the sampled object pose if it is not reachable by the robot.

\input{figures/spatial_rels}
\input{figures/mainbenchmark_time}
\subsection{Extending Spatial Relationships}\label{sec:spatial_relationship}
Scene synthesizer~\cite{scenesynthesizer} supports placing objects randomly or placing an object side-by-side with another object (called \texttt{connect}). We extended the framework to support more spatial relationships. Figure~\ref{fig:spatial_rels} shows some spatial relationship examples. 

\paragraph{Object-parent object spatial relationship} The object-parent object spatial relationship is used to determine the object's base support surface. Sceniris supports two types of spatial relationships: \texttt{`on'} and \texttt{`inside'}. \texttt{`on'} will find parent object mesh surfaces without a ``roof'', and \texttt{`inside'} will find those surfaces with a ``roof'' to ensure that the object is placed inside the object. We implement these using Scene Synthesizer's original \texttt{labal\_support} function, and we add an option to fully exclude those surfaces without a ``roof'' to enable the \texttt{`inside'} case. Examples can be found in Figure~\ref{fig:spatial_rels}(f) and (g).

\paragraph{Object-object spatial relationships} Our schema of object-object spatial relationships can be mainly defined by the following parameters: \texttt{anchor object list} $\mathcal{O}$, \texttt{direction} $\mathbf{v}$, \texttt{distance} $d$, and \texttt{distance\_type} $d_t$. The anchor object list describes all the objects $\mathcal{O}=\{O^a_1, \dots, O^a_n\}$ that will be considered in the spatial relationship with the object to be added $O^t$. $n$ is the number of the anchor objects. 

\textbf{One anchor object ($n=1$)}. If there is only one anchor object, it describes the spatial relationship between $O^a$ and $O_t$. Under this scenario, $\mathbf{v}$ is available for $\{-x, x, -y, y, \text{Null}\}$, or equivalently, $\{left, right, front, back, \text{Null}\}$, following the original definition of Scene Synthesizer. We also provide an option to transform these pre-defined directions using a global or local reference frame. When $\mathbf{v}=Null$, $O^t$ can be placed at all directions w.r.t. $O^a$. $\mathbf{v}$ can also take a direction vector. In addition to $\mathbf{v}$, $d$ is also available for the one anchor object case. Together with $d_t$, the \texttt{distance\_type}, $\mathbf{d}$ decides how far $O_t$ should be away from $O_a$. $d_t$ has three options in this case, $\{\text{`greater'}, \text{`less'}, \text{`equal'}\}$, for placing object at least $d$, at most $d$, or (softly) exact $d$ far from the anchor. To represent this spatial relationship, we construct a (partial) annulus $S_r$ (approximated by a polygon) using given $\mathbf{v}$ and $d_t$, and the points inside the annulus are considered to satisfy the condition. In detail, there will be another parameter \texttt{direction\_angle\_threshold}, $\theta$, that controls the maximum angle that deviates from $\mathbf{v}$. Formally, the four key vertices constructing the annulus are $\{p(O_a)+\text{Rot}(\mathbf{v, \pm\theta})*\text{min\_r}, p(O_a)+\text{Rot}(\mathbf{v, \pm\theta})*\text{max\_r}\}$, where $p(\cdot)$ represents the 2D position of the object on the support plane, $\text{Rot}(\cdot, \theta)$ is rotating the direction vector by $\theta$ angle on the 2D plane, and $\text{min\_r}, \textit{max\_r}$ are determined by $d_t$, such that
\vspace{5pt}
\begin{equation*}
  \text{min\_r} =
    \begin{cases}
      0 & \text{if $d_t$ is `less'}\\
      d & \text{if $d_t$ is `greater'}\\
    \end{cases}
  \text{max\_r} = 
    \begin{cases}
      d & \text{if $d_t$ is `less'}\\
      +\text{inf} & \text{if $d_t$ is `greater'}\\
    \end{cases}.
\end{equation*}
\vspace{5pt}
We then update the final polygon $S'=S_r\cap S$ that object 2D positions are sampled from, where $S$ is the original complete support plane. Examples of single-anchor relationships can be found in Figure~\ref{fig:spatial_rels}(a,b,c,d).

In addition to position relationships, we also provide a `\texttt{face\_to}' attribute to force an object's orientation to face towards another object.

\textbf{Two or more anchor objects ($n\geq2$)}. If there are two or more anchor objects, the pair-wise object relationships could be ambiguous. Therefore, we only provide one type of spatial relationship, $d_t=\text{`middle'}$, meaning that the $O^t$ should be placed in the space surrounded by $\mathcal{O}$. We construct a polygon where the vertices of the polygon are the positions of anchor objects $\{p(O^a_1), ...p(O^a_n)\}$, which is used to sample points for object placement. An example of this relationship can be found in Figure~\ref{fig:spatial_rels}(e)

\paragraph{Object-support surface relationship} We add a configuration term `ratio\_on\_support' to loosely control how the object should be placed concerning the surface boundary. When the `ratio\_on\_support` is set to 1.0, the surface should ideally hold the entire object's projection, i.e., $P(\text{obj}) \subset \text{surface}$, where $P(\cdot)$ is the projection of the object mesh to the support plane. When the value is set to 0, it is a valid placement as long as the pivot (usually the center of the bottom surface of the object's bounding box) is on the surface. To ensure efficiency, we implement this through estimation ahead of sampling the actual poses of the objects. We use the ratio\_on\_support $\times$ half length of the shorter edge of the projected object mesh to erode the support surface. So there will be cases where the object does not exactly follow the configured ratio\_on\_support value.

\paragraph{Other improvements} We improved other engineering aspects of the original Scene Synthesizer~\cite{scenesynthesizer}. We cache the asset-level Trimesh scene, the result of \texttt{asset.as\_trimesh\_scene()}, because we find the operation is time-consuming, and it is used in a lot of places. We modified the visualization so that the generated batch of scenes can be visualized in the same window.

\input{figures/mainbenchmark_time_breakdown}
\section{EXPERIMENTS}
\subsection{Batched Scene Generation}\label{sec:batch_scene_generation}
We first benchmark the scene generation speed of Sceniris and Scene Synthesizer using a scene (Figure~\ref{fig:mainbenchmark_time}(a)) containing three objects: an apple, a banana, and a cabinet. The apple and the cabinet will be randomly placed on the plane, and the banana will be placed inside the upper drawer of the cabinet. The upper and lower drawer joints should be in random states. This scene configuration only contains spatial relationships that are supported in both Sceniris and Scene Synthesizer, so we can perform an apples-to-apples comparison. The benchmark is conducted on a Google Cloud VM with 16 CPU threads (Intel Xeon CPU @ 2.20GHz) and an NVIDIA L4 GPU. Since Scene Synthesizer operates on a single thread, we run Scene Synthesizer~\cite{scenesynthesizer} on 10 processes simultaneously, and we evenly distribute the scene instances to generate for those processes. Each object is allowed 10 retries when a collision is detected. If a scene instance fails to meet the collision-free condition, the instance will be marked invalid.

Figure~\ref{fig:mainbenchmark_time} shows the system-wise comparison regarding the execution time (b) and the number of valid scenes generated per second (c), between Sceniris and Scene Synthesizer. We vary the total number of scenes to generate from 4 to 16,384. On the execution time, Sceniris can finish the operation on 16,384 scenes in 32s, while Scene Synthesizer can finish 64 scenes in 39s. Considering the number of scenes they operate on, Sceniris achieves about 311 times the efficiency of Scene Synthesizer. This shows Sceniris is a highly efficient system for generating a large batch of random instances for the same configuration. Sceniris achieves another significant efficiency improvement when switching to a warm start case, i.e., only executing the "Generation" step in Figure~\ref{fig:main_steps}. The warm started Sceniris finishes 16,384 scenes in 2.52s, even shorter than Scene Synthesizer operating on 4 scenes. This shows that we significantly reduced the repeated overhead cost when re-sampling the entire scene. Therefore, Sceniris is not only able to generate a batch of scene instances efficiently, but is also able to continuously generate randomized instances, which is promising in parallel reinforcement learning or data collection scenarios.
In Figure~\ref{fig:mainbenchmark_time}(c), we show that Sceniris generates about 234 times more valid scenes than Scene Synthesizer in the cold start case and about 2,936 times more in the warm start case. 

\vspace{-5pt}
\subsection{Component Breakdown}
We profile the execution time of each key step in the generation procedure in Sceniris and Scene Synthesizer, using the same scene configuration as above. 
According to Figure~\ref{fig:mainbenchmark_time_breakdown}, Sceniris spends 10.17s on generating \textbf{1,024} scenes and Scene Synthesizer takes 2.99s for \textbf{1} scene. Sceniris spends significantly less portion of execution time on placing objects (green) compared to Scene Synthesizer, showing that our successful effort on minimizing the recurrent overhead cost. In comparison, Scene Synthesizer spends a huge portion of time on overhead other than collision checking for objects, shown by ``other overhead'' on the left of Figure~\ref{fig:mainbenchmark_time_breakdown}. 
We later find that such overhead in Scene Synthesizer is caused by dumping the object asset to a trimesh scene repeatedly. 
We then, in Sceniris, implemented a cache mechanism to perform this operation only once. 
We also perform the benchmark for Sceniris with the reachability check turned on for the apple and the cabinet. We find that the running time is almost the same without the reachability check. Our reachability check only takes a constant $\sim$0.0001s per query, which is at most 1/20 of the collision check cost.

Sceniris spends a significant amount of time on initializing the collision checker, which we use cuRobo to implement. In the pressure test below, we find that the time for making cuRobo world configurations increases with the number of scenes and some overhead is presented. 
We believe this overhead will be improved by a better cuRobo interface that improves the creation of world configurations, which is out of scope of this work and will be left for future work. 

\vspace{-10pt}
\subsection{Pressure Test} We conduct a simple pressure test using the same scene configuration in batch scene generation experiment. 
We find that generating a batch of 524,288 scenes fits in an L4 GPU, taking about 20GB VRAM. It takes 769.67s to finish for the cold start case and has 113,820 invalid scenes, resulting in 533.30 valid scenes/s. This is higher than 408.77 valid scenes/s in Figure~\ref{fig:mainbenchmark_time}, showing that our cold start performance is not saturated, and running on a larger batch and a GPU with larger VRAM could continuously improve this performance. For the warm start performance, it achieves 5,279.84 valid scenes/s.

\vspace{-10pt}
\subsection{Scaling up with Complex Spatial Relationships}
\input{figures/cfg_levels}
Based on the standard benchmark scene configuration above (named Mid), we extend harder scene configurations (Hard, Hard+) with our complex spatial relationships and test their generation time. In the Hard scene configuration, we keep the drawer and the banana from mid, and place the apple in front of the drawer at least 0.5 meters away. In addition, the orange should be placed between the drawer and the apple. In Hard+ configuration, we add a mug that should be placed to the right of the drawer within a 0.7m range. The orange should be placed between the drawer, the mug, and the apple. Hard contains 2 complex spatial relationships, and Hard+ contains 3, and the "in between" is harder since there is one additional object involved. In this experiment, we are not able to compare with Scene Synthesizer because it does not support these spatial relationships.

We report the execution speed in Figure~\ref{fig:cfg_levels}. We observe that Hard and Hard+ increase about 20-30s execution time over its previous level in the cold start case, and about 20s in the warm start case. 
This is because some of the complex spatial relationships can not be propagated to the batch by a simple rigid transformation, and Sceniris has to compute the valid sample polygon per scene instance. 
Nevertheless, Sceniris still achieves about 249 valid scenes/s in Hard+ (cold start), showing that Sceniris has the efficiency advantage compared to Scene Synthesizer. We believe this could be sped up by batch sampling from heterogeneous polygons on a GPU, which is a potential future work.

\input{figures/example_scenes}
\vspace{-12pt}
\subsection{Qualitative results}
In Figure~\ref{fig:example_scenes}, we demonstrate a kitchen scene and a table top scene generated from Sceniris. The scene includes a table, a chair next to the table, a cabinet with two drawers placed on the table, a mug next to the cabinet, and a banana placed on the top of the cabinet. We highlight our capability of maintaining object spatial relationships and randomizing joint states.

\vspace{-7pt}
\section{CONCLUSIONS}
We present Sceniris, a fast procedural scene generation framework using the ideas of parallelization and GPU computing. Sceniris can generate collision-free and robot-reachable scenes that provide rich support for physical AI, learning based scene generation, and 3D understanding applications. Sceniris achieves 200+$\times$ speed-up over the naively multi-processed base approach Scene Synthesizer~\cite{scenesynthesizer}.








\bibliography{main}
\bibliographystyle{ieeetr}

\end{document}

%% file: figures/teaser.tex

\begin{center}
\vspace{-20pt}
\includegraphics[width=0.95\textwidth]{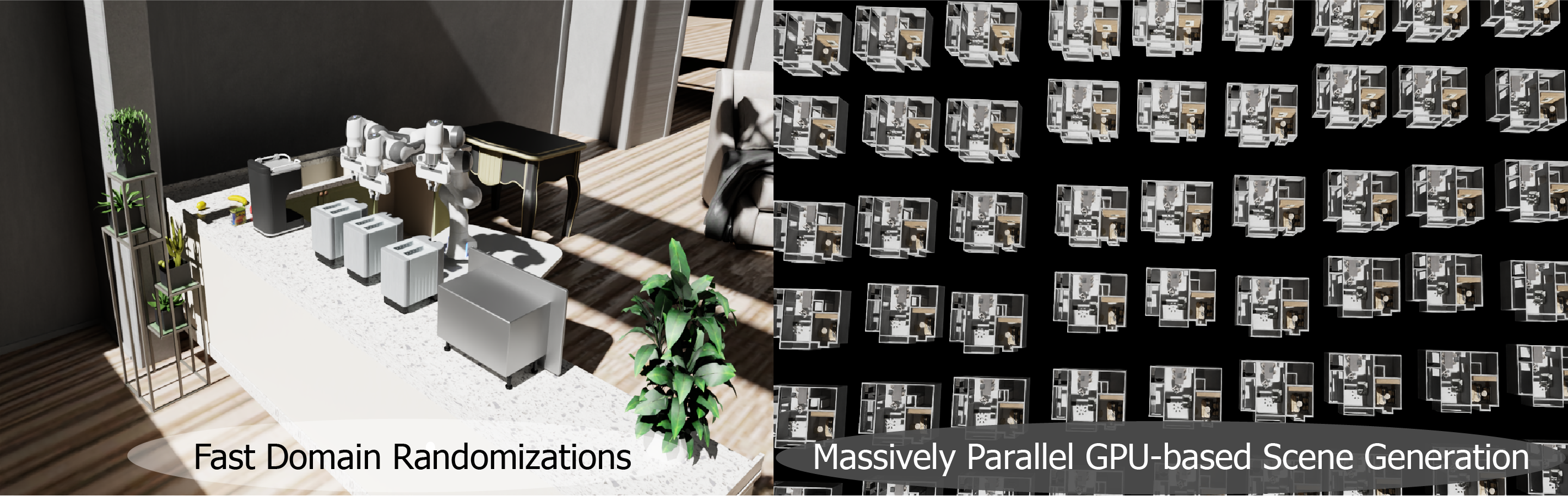}
    \captionof{figure}{Sceniris is a procedural scene generation tool that provides fast, scalable, and physical-AI plausible scene generation. We leverage GPU to accelerate the collision checking and robot reachability check.}
    \label{fig:placeholder}
\end{center}

%% file: figures/main_steps.tex
\begin{figure*}[tbh]
    \centering
    \vspace{-25pt}
    \includegraphics[width=\textwidth]{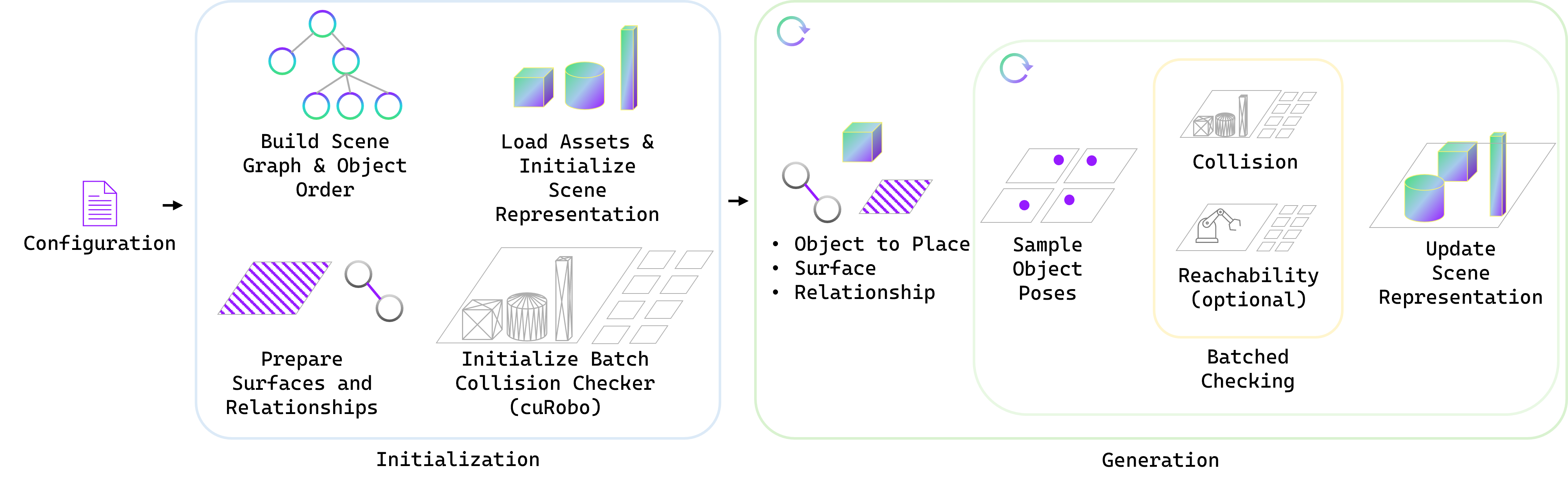}
    \vspace{-20pt}
    \caption{Sceniris takes a configuration and generates a batch of scenes efficiently. Alternatively, Sceniris can generate a scene by adding objects step-by-step, but the efficiency is not optimized.}
    \label{fig:main_steps}
    \vspace{-10pt}
\end{figure*}

%% file: tables/sg_comparision.tex
\begin{table*}[thb]
    \centering
    \setlength{\tabcolsep}{3pt}
    \caption{Comparison between Sceniris and existing procedural scene generation approaches, including standalone approaches and embedded algorithms in simulation frameworks}
    \vspace{-3pt}
    \resizebox{\textwidth}{!}{
        \begin{tabular}{ccccccc}
        \toprule
        Framework & Scene Representation & Sampling & Collision Checking & Batched & Spatial Relationships & Reachability \\
        \midrule
        \hspace{-78pt}\textit{\textbf{Standalone}} \\
        Sceniris (Ours) & Batched Trimesh & Rejection Sampling & cuRobo & yes & pair and multi-object & \cmark \\
        Scene Synthesizer~\cite{scenesynthesizer} & Trimesh & Rejection Sampling & FCL (Trimesh) & \xmark & `on', object connect  & \xmark \\
        Infinigen-indoors~\cite{raistrick2024infinigen} & Trimesh & Monte-Carlo Metropolis–Hastings & \cmark & \xmark & \cmark & \xmark \\
        Spatial Scene Grammars~\cite{izatt2020generative} & Tree & unknown & Drake & \xmark & \cmark & \xmark \\

        \midrule
        \hspace{-34pt}\textit{\textbf{In generative simulation}} \\
        RoboGen~\cite{robogen} & N/A & Rejection Sampling/Collision Resolving & PyBullet & \xmark & \{`on', `in'\} & \xmark \\
        GenSim2~\cite{gensim2} & N/A & Predefined + Small-range Sampling  & \xmark & \xmark & \xmark & \xmark \\
        GenSim~\cite{gensim} & N/A & Rejection Sampling & 2D occupancy & \xmark & \xmark & \xmark \\
        RoboTwin 2.0~\cite{robotwin2} & N/A & Small-range Sampling & \xmark & \xmark & \xmark & \xmark \\

        \midrule
        \hspace{-8pt}\textit{\textbf{In simulation and game engines}} \\
        Robocasa~\cite{robocasa} & N/A & \xmark & \xmark & \xmark & \xmark & \xmark \\
        LIBERO~\cite{libero} & BDDL & Rejection Sampling & N/A & \xmark & \xmark & \xmark \\
        VIMA~\cite{vima} & N/A & Rejection Sampling & 2D Occupancy & \xmark & \xmark & \xmark \\
        RLBench~\cite{rlbench} & CoppeliaSim & Rejection Sampling & Bounding Box & \xmark & \xmark & \xmark \\
        ProcTHOR~\cite{procthor} & Trimesh & Rejection Sampling & FCL (Trimesh) & \xmark & \cmark & \xmark \\
        UE5 PCG~\cite{ue5pcg} & Graph & Rejection sampling & Built-in Tool & \xmark & \cmark & \xmark \\
        \bottomrule
        \end{tabular}
    }
    \vspace{-10pt}
    \label{tab:sg_comparision}
\end{table*}

%% file: figures/spatial_rels.tex
\begin{figure}[htb]
    \centering
    \vspace{-10pt}
    \includegraphics[width=0.5\textwidth]{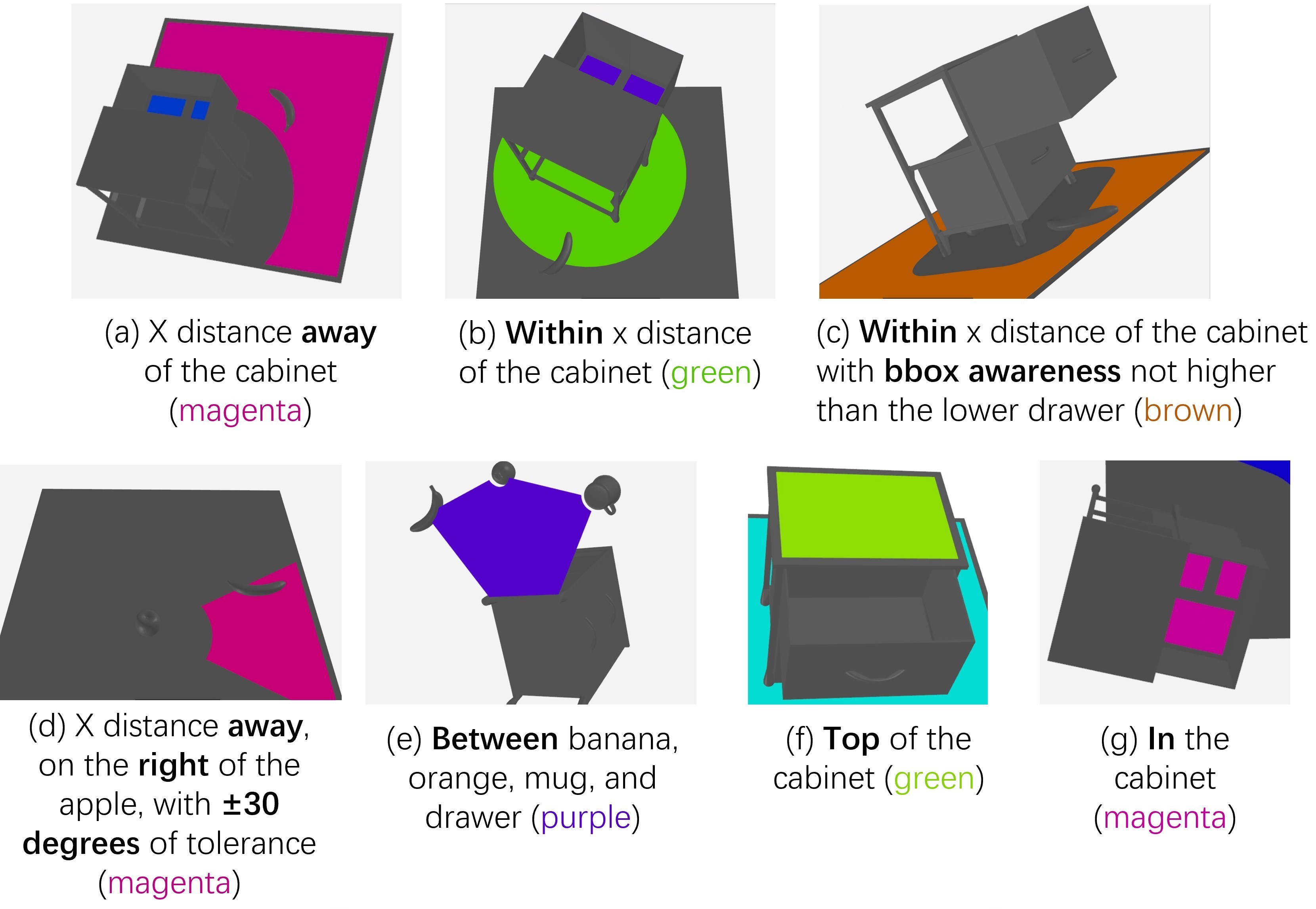}
    \vspace{-20pt}
    \caption{Examples of additional spatial relationships supported by Sceniris}
    \vspace{-20pt}
    \label{fig:spatial_rels}
\end{figure}

%% file: figures/mainbenchmark_time.tex
\begin{figure*}[tb]
    \centering
    \vspace{-25pt}    
    \includegraphics[width=0.9\textwidth]{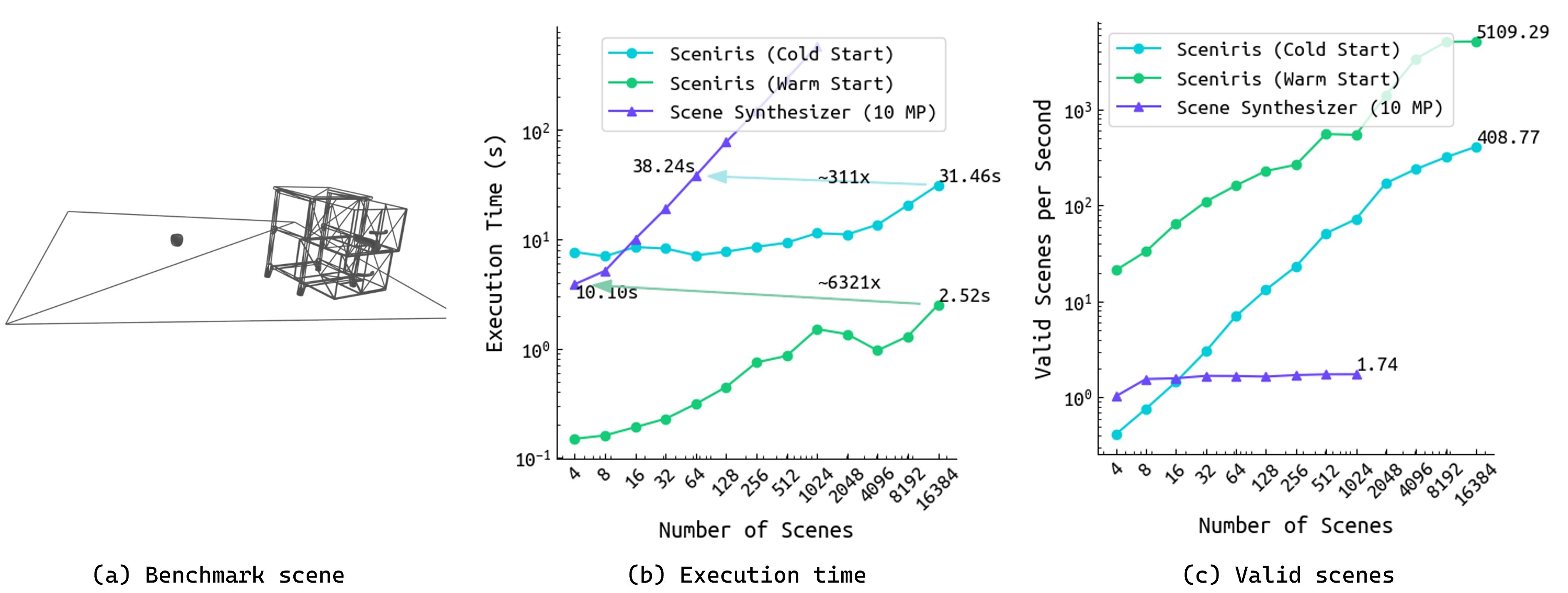}
    \vspace{-10pt}
    \caption{(a) The benchmark scene, simplified for visualization. (b) Execution time. (c) The number of valid environments generated per second. Sceniris handles 311 and 6,321 times more environments per second than 10 multi-processed Scene Synthesizer, under cold start and warm start, respectively.}
    \label{fig:mainbenchmark_time}
    \vspace{-15pt}
\end{figure*}

%% file: figures/mainbenchmark_time_breakdown.tex
\begin{figure*}[tb]
    \centering
    \vspace{-20pt}
    \includegraphics[width=0.9\textwidth]{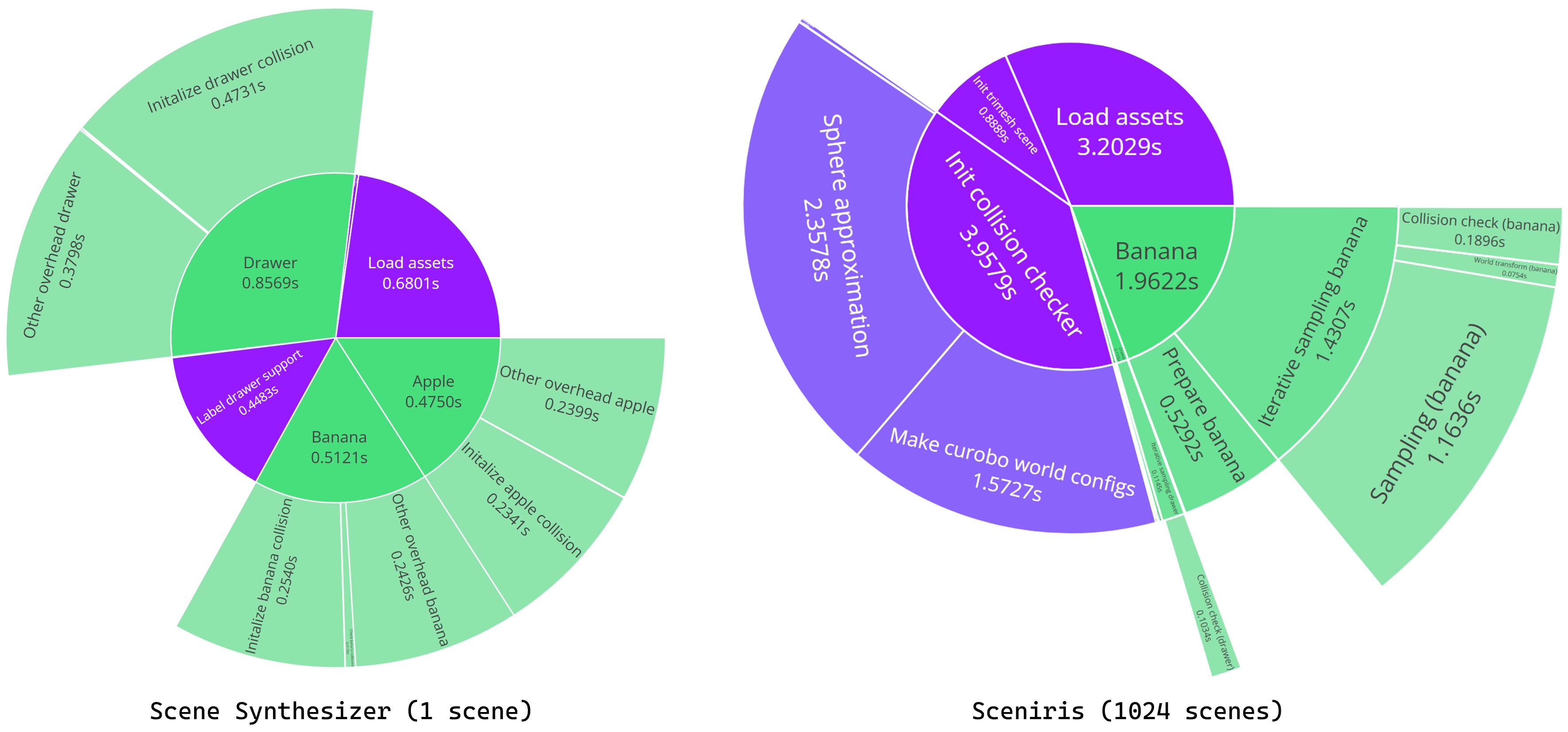}
    \vspace{-10pt}
    \caption{Total time spent on key steps in Scene Synthesizer (\textbf{1} scene) and Sceniris (ours, \textbf{1024} scenes). Both approaches use the same scene configuration in Section~\ref{sec:batch_scene_generation}. Regions in green show the time spent on placing objects.}
    \vspace{-10pt}
    \label{fig:mainbenchmark_time_breakdown}
\end{figure*}

%% file: figures/cfg_levels.tex
\begin{figure}[h]
    \centering
    \vspace{-10pt}
    \includegraphics[width=0.30\textwidth]{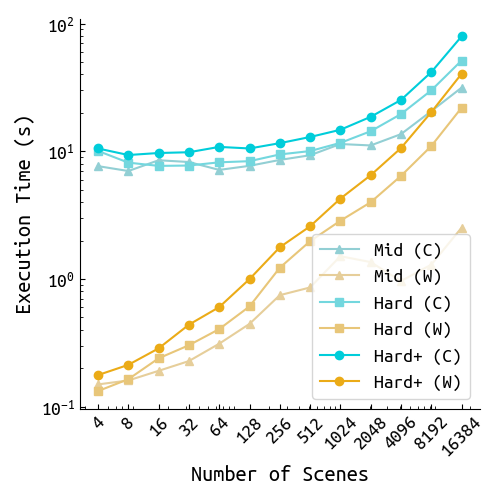}
    \vspace{-10pt}
    \caption{Execution time of different scene configurations. C: Cold start; W: Warm start. Mid is the basic scene configuration. Hard has 2 complex spatial relationships and Hard+ has 3.}
    \label{fig:cfg_levels}
    \vspace{-10pt}
\end{figure}

%% file: figures/example_scenes.tex
\begin{figure}[tbh]
    \centering
    \vspace{-5pt}
    \includegraphics[width=\linewidth]{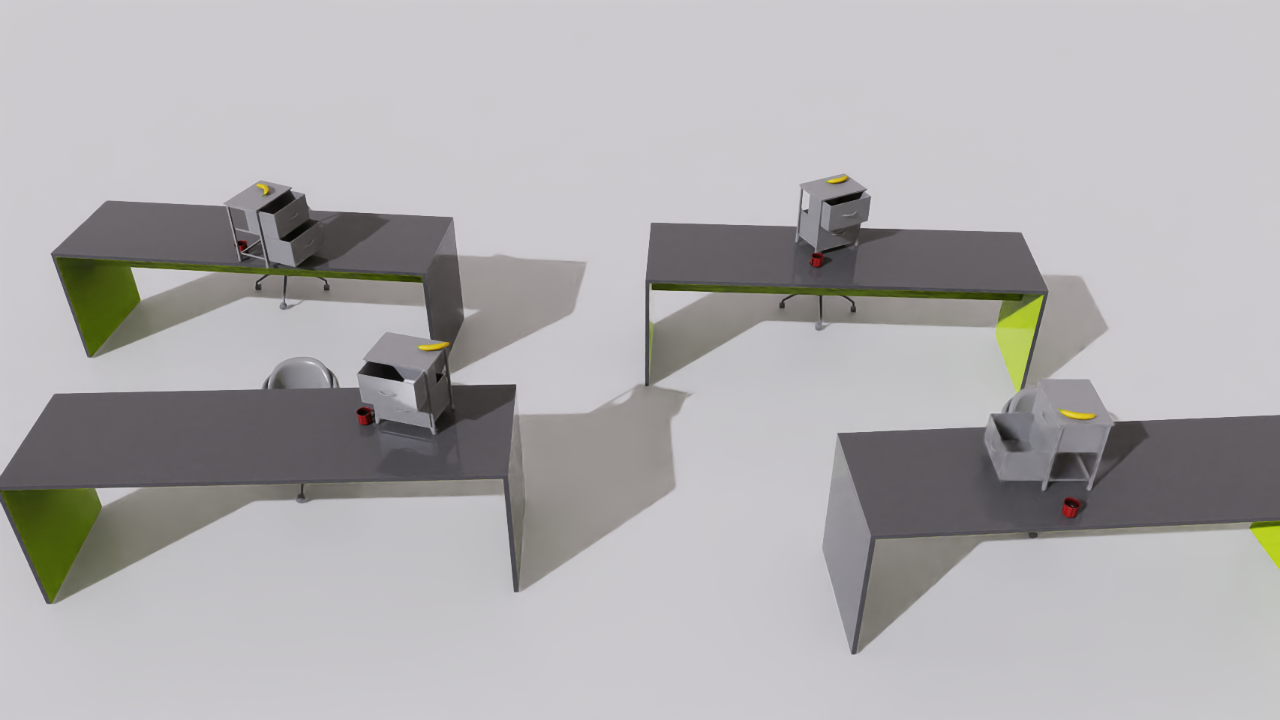}
    \vspace{-10pt}
    \caption{Example of generated scenes.}
    \label{fig:example_scenes}
\end{figure}